\documentclass[11pt,a4paper]{article}
\usepackage[hyperref]{acl2020}
\usepackage{times}
\usepackage{latexsym}

\usepackage{microtype}

\aclfinalcopy %

\usepackage{microtype}
\usepackage{url}
\usepackage{times}  %
\usepackage{helvet}  %
\usepackage{courier}  %
\usepackage{url}  %
\usepackage{graphicx}  %

\usepackage{soul}
\usepackage[utf8]{inputenc}
\usepackage[]{caption}
\usepackage{multirow}
\usepackage{amsmath}
\usepackage{amsthm}
\usepackage{amssymb}

\usepackage{latexsym}
\usepackage{subcaption}
\usepackage{CJKutf8} 
\usepackage{color}
\usepackage{bm}
\usepackage{booktabs}
\usepackage{threeparttable}
\usepackage[colorinlistoftodos]{todonotes}
\newcommand{\ttrule}{\specialrule{.05em}{0pt} {.65ex}}
\newcommand{\ddrule}{\specialrule{.05em}{.4ex}{0pt}}

\title{End-to-End Chinese Parsing Exploiting Lexicons}

\author{First Author \\
  Affiliation / Address line 1 \\
  Affiliation / Address line 2 \\
  Affiliation / Address line 3 \\
  \texttt{email@domain} \\\And
  Second Author \\
  Affiliation / Address line 1 \\
  Affiliation / Address line 2 \\
  Affiliation / Address line 3 \\
  \texttt{email@domain} \\}

\author{Yuan~Zhang, Zhiyang~Teng, Yue~Zhang\\
	School of Engineering, Westlake University, China \\
	Institute of Advanced Technology, Westlake Institute for Advanced Study\\
	\texttt{chuengyuenn@gmail.com}\\
	\texttt{\{tengzhiyang,zhangyue\}@westlake.edu.cn}\\}

\date{}

\begin{document}
\begin{CJK*}{UTF8}{gbsn}
\maketitle

\begin{abstract}
Chinese parsing has traditionally been solved by three pipeline systems including word-segmentation, part-of-speech tagging and dependency parsing modules. 
In this paper, we propose an end-to-end Chinese parsing model based on \emph{character inputs} which jointly learns to output word segmentation, part-of-speech tags and dependency structures. 
In particular, our parsing model relies on word-char graph attention networks, which can enrich the character inputs with \emph{external word knowledge}. 
Experiments on three Chinese parsing benchmark datasets show the effectiveness of our models, achieving the state-of-the-art results on end-to-end Chinese parsing. 
\end{abstract}
\section{Introduction}

\begin{figure}[!tb]  
 \setlength{\abovecaptionskip}{6pt}
 \setlength{\belowcaptionskip}{-8pt}
   \centering
   \includegraphics[width=1\linewidth]{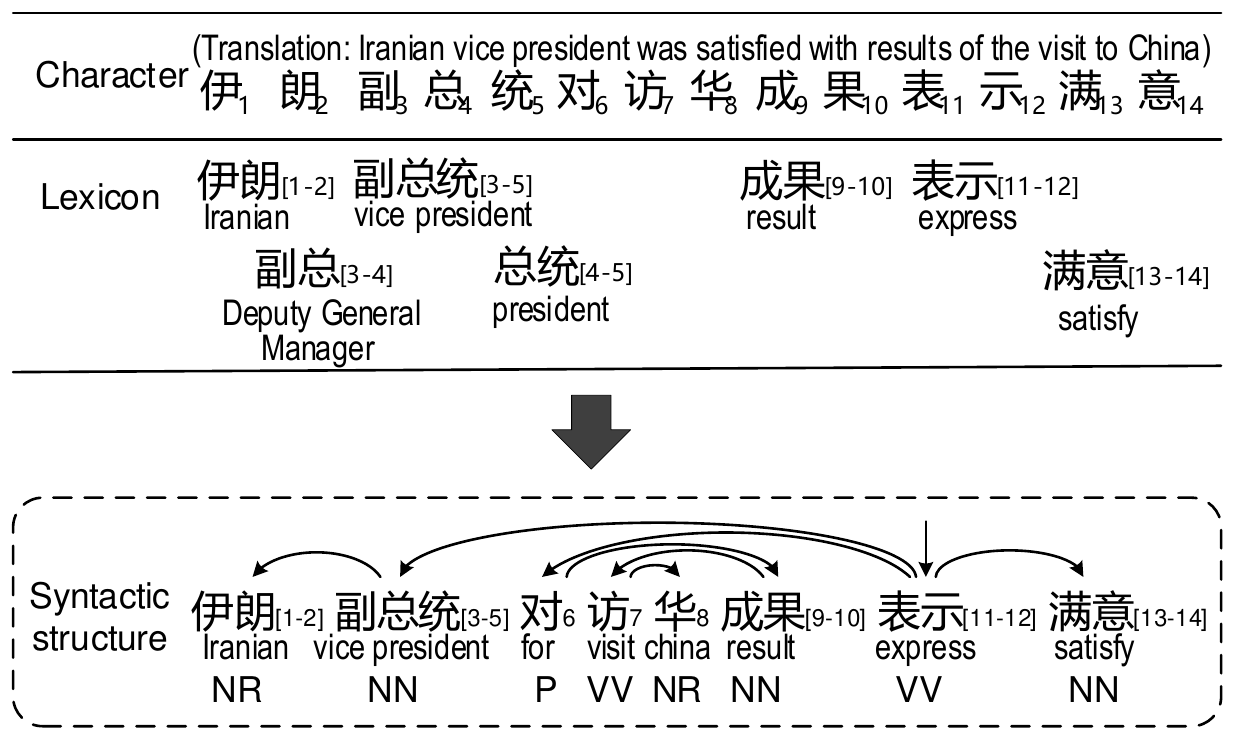}
   \caption{End-to-end parsing exploiting lexicons.}
      \label{fig:end2endparse}
\end{figure}

As a fundamental task in syntactic analysis, dependent parsing has received constant research attention \citep{dozat2016deep,  dozat-etal-2017-stanfords, strubell-etal-2018-linguistically,ma2018stack, li2019self}. It offers useful information to a range of downstream tasks, such as relation extraction \citep{gamallo2012dependency, miwa-bansal-2016-end, guo-etal-2019-attention, zhang-etal-2018-graph} and semantic parsing \citep{hajic-etal-2009-conll,poon-domingos-2009-unsupervised,sun-etal-2018-semantic}. As shown in Figure \ref{fig:end2endparse}, the goal of syntactic dependency parsing is to build a dependency tree for a given sentence, where each arc represents a head-dependent relationship between two words.

Traditionally, Chinese dependency parsing takes word segmentation and POS tagging as preprocessing steps \citep{zhou2000block, ma2012fourth, zhangmcdonald2014enforcing}. 
The pipeline method, however, suffers from error propagation as incorrect word boundaries and POS tags lead to decreases in parsing performance. 
End-to-end models, which take character sequences as input and jointly perform the three task, have been investigated to address the problem \citep{hatori2012incremental, zhang2013chinese, kurita2017neural, li2018neural}.

While most existing work takes a transition-based method \citep{ Nivre:2008:CL,zhangnivre2011transition,bohnet:2010:coling,chen2014dep,andor:2016:global}, 
we consider a graph-based method for end-to-end parsing, adopting the bi-affine framework of  \citet{dozat2016deep}. 
In particular, 
our model takes a character sequence as input, using a sequence representation network to find the  representation of each character. Word segmentation, POS-tagging and parsing are performed jointly over the character representation by multi-task learning. 
More specifically, both word segmentation and POS-tagging are performed as character sequence labeling tasks, where a local classifier is built on top of each character representation. Dependency parsing follows a bi-affine scoring function between characters, so that the head of each character can be found.
Both bidirectional long short-term memory networks (BiLSTMs) and self attention networks (SANs;  \citeauthor{vaswani2017attention}, \citeyear{vaswani2017attention}) are considered as the encoder network.

One salient difference of neural graph-based parsing, as compared with its transition-based counterpart, is that the representation of input is calculated first, before local outputs such as sequence labels and bi-word relation tags are predicted. In contrast, a transition-based parser builds a joint output structure using a state-transition process, the representation of which contains mixed input and output information. As a result, one advantage of neural graph-based parsing is that the representation calculation can be more parallelizable, allowing faster running speed. However, for our joint parser, since word segmentation and dependency parsing are performed jointly on characters, word information cannot be used directly to benefit parser disambiguation. 

To solve this issue, we consider integrating lexicon knowledge for enriching the character sequence representation, by jointly encoding the characters and all the words in the input that match a dictionary. To this end, we use lattice LSTM \citep{zhang-etal-2019-lattice} for extending a BiLSTM character encoder, and make a novel extension to the Transformer architecture for the SAN counterpart. The latter runs over a order of magnitude faster than the former thanks to strong parallelization.
In particular, the Transformer \citep{vaswani2017attention} architecture is adopted, which can be regarded as a graph-attentional neural network \citep{velikovi2017graph} with a fully-connected character graph. We integrate information from lexicon words into this graph attentional neural network by taking them as additional vertices in the graph, adding word-character edges and word-word edges to the input graph. The standard self-attention function is further extended into a novel combination of  a semantic channel and a structural channel, 
the former using semantic similarity for weight calculation, and the latter taking paths in the graph into consideration.

Experiments on three Chinese parsing datasets show that 
integrating lexicon word information is useful for improving character-level end-to-end parsing. 
Our graph-based parser, enriched with word-level features, outperforms all existing methods in all the datasets, achieving the best results on segmentation, POS tagging and parsing in the literature.  To our knowledge, we are the first to investigate an end-to-end Chinese parser exploiting lexicon knowledge.

\section{Related Work}
\paragraph{End-to-End Chinese Parsing.}  
\citet{hatori2012incremental} pioneer research on the joint model of  word segmentation, POS tagging, and dependency parsing for Chinese using transition-based methods.  \citet{zhang2014character} exploit the manually annotated  intra-character dependencies. \citet{zhang2015randomized} consider transition-based joint word segmentation, POS tagging and dependency reranking using randomized greedy inference. \citet{kurita2017neural} first investigate transition-based models for joint Chinese lexical and syntactic analysis with neural models. We investigate the same task, but our end-to-end models are built on graph-based parsers. 

Very recently, \citet{yan2019unified} consider using a BiLSTM and BERT encoder for representing character sequences for end-to-end Chinese parsing. Our work is similar in being a graph-based parser, but differs in three aspects. First, we investigate the effectiveness of word information for the task by considering a novel graph attention network with semantic and structural channels. Second, we compare BiLSTM encoding with Transformer encoding with BERT. Third, while they consider joint segmentation and parsing, we consider joint segmentation, POS-tagging and parsing.

\paragraph{Word-character Lattice Neural Networks.} 
\citet{chen2017dag} and \citet{zhang-yang-2018-chinese} use lattice LSTM to deal with mixed word and character inputs. 
\citet{ding-etal-2019-neural} use graph convolutional networks \citep{kipf2016semi} for entity lattice inputs. Their lattice requires named entities and their entity types as inputs. More closely related to our work, \citet{sperber-etal-2019-self} adapt Transformer \citep{vaswani2017attention} for lattice inputs. Our models are different from them in three aspects. First, their lattice is built from compressing speech hypothesis, while we build the lattice by lexicon matching. Second, there is no concept of word in their models, while we explicitly model word inputs and exploit pretrained word embeddings. Third, we differentiate semantic and structural information according to the edge types in lattice by taking inspirations from graph Transformers \citep{velikovi2017graph,zhu-etal-2019-modeling}. Existing work can be regarded as a variant of our models, which only captures semantic features.

\section{Task Description}

Formally, as shown in Figure \ref{fig:end2endparse}, given a sequence of characters $s=c_1, ..., c_n$, 
the target of end-to-end parsing is to obtain a dependency tree $T=(V,E)$ together with
segmented word sequence $s_w={w_1, ..., w_k}$ and a 
POS tag sequence $s_t=t_1, ..., t_k$, 
where $c_i$ is the $i$-th Chinese character in the sentence, $w_j$ is the $j$-th segmented word and $t_j$ is the POS tag for $w_j$.  
The node set $V$ contains all segmented words and an additional root dummy node. 
The arc set $E=\left \{ (i_k, j_k) \right \}$, where the tuple $(i_k, j_k)$ represents $w_{i_k}$ is the head of $w_{j_k}$.

\section{LSTM Encoder}

\begin{figure*}[!tb]  
  \centering
  \includegraphics[width=1\linewidth]{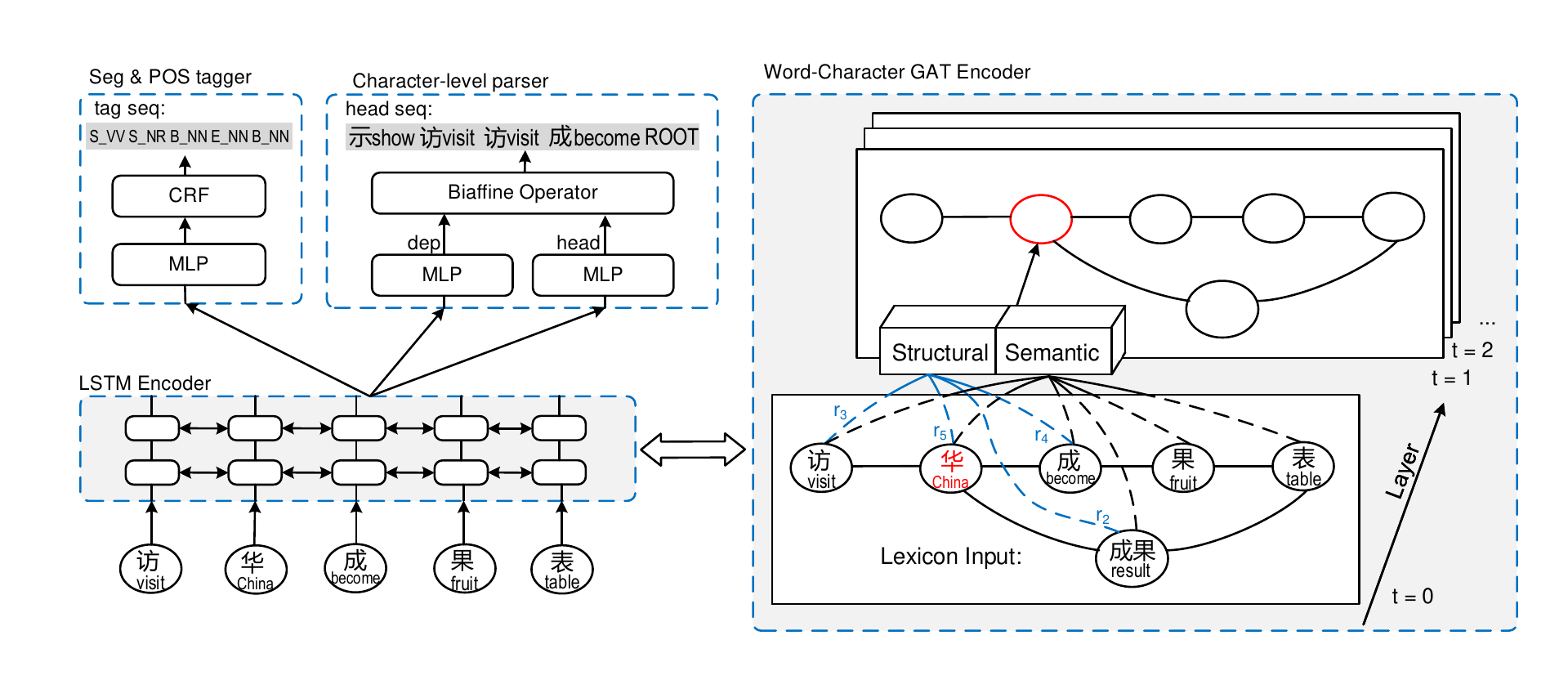}
  \caption{End-to-end Chinese parsing. The input is a part of the sentence in Figure \ref{fig:end2endparse} with character index $i \in [7, 11]$. The output sequences are the gold labels accordingly. The output structure for the entire sentence is shown in Figure \ref{fig:end2endparse}. }
      \label{fig:biaffine}
\end{figure*}

The overall framework of our method is shown in Figure \ref{fig:biaffine}. 
In particular, we adopt graph-based parsing and use biaffine transformation for dependency arc prediction. For word segmentation and POS tagging,  a  CRF is used for predicting label sequences.
We use recurrent neural networks for input representations, following \citet{dozat2016deep}, \citet{kiperwasser2016simple} and \citet{wang-chang-2016-graph}. Specifically, we obtain character embeddings through BERT: 
$\mathbf{e}_{c_i}=\text{emb}(c_i)$.

A multi-layer bi-directional LSTM structure is used to calculate the character sequence representations. In particular, the initial character embedding $\mathbf{e}_{c_i}$ are denoted as $\mathbf{h}^0_{c_i}$. Subsequently, for the $k$-th layer, $\mathbf{h}^{k}_{c_1}, ..., \mathbf{h}^{k}_{c_n}$  are calculated from $\mathbf{h}^{k-1}_{c_1}, ..., \mathbf{h}^{k-1}_{c_n}$ as follows:
 \begin{flalign*}
\overrightarrow{\mathbf{h}^k_i}, \overrightarrow{\mathbf{c}^k_i} &= \overrightarrow{\textit{LSTM}^k} (\overrightarrow{\mathbf{h}^{k-1}_i}, \overrightarrow{\mathbf{h}^k_{i-1}}, \overrightarrow{\mathbf{c}^k_{i-1}}), 
\end{flalign*}
 \begin{flalign*}
\overleftarrow{\mathbf{h}^k_i}, 
\overleftarrow{\mathbf{c}^k_i} &= \overleftarrow{\textit{LSTM}^k} (\overleftarrow{\mathbf{h}^{k-1}_i}, \overleftarrow{\mathbf{h}^k_{i+1}}, \overleftarrow{\mathbf{c}^k_{i+1}}), \\
\mathbf{h}^k_i &= \langle \overleftarrow{\mathbf{h}^k_i}, \overrightarrow{\mathbf{h}^k_i} \rangle, \ \ \ \ 
\mathbf{c}^k_i = \langle \overleftarrow{\mathbf{c}^k_i}, \overrightarrow{\mathbf{c}^k_i} \rangle,
\end{flalign*}
where $\textit{LSTM}^k$ is the LSTM network for the $k$-th layer,  $\mathbf{h}^k_i$ and $\mathbf{c}^k_i$ are the hidden output and the internal cell state of the $i$-th character. $\rightarrow$ and  $\leftarrow$ indicate the left and right composition directions of LSTM networks respectively. The final representation of the $i$-th character $c_i$ is the output hidden vector of the $K$-th layer: $\mathbf{h}_{c_i}=\mathbf{h}^K_{c_i}$.

We follow the lattice LSTM extension \citep{zhang-etal-2019-lattice} for integrating word features from a dictionary. Formally, the hidden states are calculated as follows:

 \begin{flalign*}
\overrightarrow{\mathbf{h}^k_i}, \overrightarrow{\mathbf{c}^k_i} &= \overrightarrow{\textit{LatticeLSTM}^k} (\overrightarrow{\mathbf{h}^{k-1}_i}, \overrightarrow{\mathbf{h}^k_{i-1}}, \overrightarrow{\mathbf{c}^k_{i-1}}, \mathbf{E}_{w_{\cdot i}} ),\\
\overleftarrow{\mathbf{h}^k_i}, \overleftarrow{\mathbf{c}^k_i} &= \overleftarrow{\textit{LatticeLSTM}^k} (\overleftarrow{\mathbf{h}^{k-1}_i}, \overleftarrow{\mathbf{h}^k_{i+1}}, \overleftarrow{\mathbf{c}^k_{i+1}}, \mathbf{E}_{w_{i \cdot}} ),
\end{flalign*}
 \begin{flalign*}
\mathbf{h}^k_i &= \langle \overleftarrow{\mathbf{h}^k_i}, \overrightarrow{\mathbf{h}^k_i} \rangle, \ \ \ \ 
\mathbf{c}^k_i = \langle \overleftarrow{\mathbf{c}^k_i}, \overrightarrow{\mathbf{c}^k_i} \rangle,
\end{flalign*}
where $\textit{LatticeLSTM}^k$ is the Lattice-LSTM network for the $k$-th layer. $\mathbf{E}_{w_{\cdot i}}$ and $\mathbf{E}_{w_{i \cdot}}$ are the sets of word embeddings for words ending and starting with the $i$-th character, respectively.

\section{Word-Character GAT Encoder}
\label{sec:sub:latticeatt}

We extend the standard transformer \citep{vaswani2017attention}. As shown in Figure \ref{fig:biaffine}, the word-character GAT encoder is used as one alternative encoder for the LSTM encoder component in the earlier section to model a word-char mixed sequence. The model consists of a multi-layer encoder. Each layer consists of two sub-layers, including a multi-head self-attention sublayer and a position-wise feed-forward network sublayer. Layer normalization and residual network are used for each sublayer. 
Formally, denote the input character-word mixed sequence as $t=[c_1, ..., c_n, w_1, ..., w_k]$, 
where $c_1, ..., c_n$ represent the input character sequence and $w_1, ..., w_k$ represent the words in the input that match a lexicon. For the convenience of notation, we represent each token in the input as $t_i (i \in [1, n+k])$, regardless of whether it is a character ($i<n+1$) or word ($i>n$).

\textbf{Position Embedding} We use static position embeddings to encode the input following \citet{vaswani2017attention}. 
$e^p(i)$ represents the position embedding for the $i$-th position. 
We inject position embedding to characters and word input as follows:
\begin{flalign*}
\mathbf{e}'_{c_i}=\mathbf{e}_{c_i}+\mathbf{e}^p(i); 
\mathbf{e}'_{w_i}=\mathbf{e}'_{w_i} + \mathbf{e}^p(b_{w_i}).
\end{flalign*}

We take the position of the first character $b_{w_i}$ of the word $w_i$ in the original sentence to represent the position of the word.

\begin{table}[!t]
\small
\centering
\begin{tabular}{ccll}
\ttrule
\textbf{Block} & & \textbf{Relation} & \textbf{Example}    \\ 
\midrule
\multirow{5}{*}{\shortstack{A\\ (char)}} & 1 & word $\rightarrow$ char & - \\
& 2 & char $\leftarrow$ word &  成果(result):华(China)   \\
& 3 & char $\rightarrow$ char &  访(visit):华(China)   \\
& 4 & char $\leftarrow$ char &   成(become):华(China)  \\
& 5 & self-to-self & 华(China):华(China)   \\ 
\midrule
\multirow{3}{*}{\shortstack{B\\ (word)}}& 6 & char $\rightarrow$ word & 华(China):成果(result) \\
& 7 & word $\leftarrow$ char &  表(table):成果(result)   \\
& 8 & self-to-self & 成果(result):成果(result)   \\ 
\ddrule
\end{tabular}
\caption{Relation definitions of word-character interaction. Relation expression ``A $\rightarrow$ B'' indicates a relation from A (left) to B (right). Block A (to a character) and Block B (to a word) take the character ``华(China)'' and word ``成果(Result)'' as examples, respectively.}
\label{tab:rels}
\end{table}

\textbf{Information Integration}
For the convenience of describing both word and characters, we unify them as nodes in a graph represented using a graph attention network \cite{velikovi2017graph}.
We define two channels for capturing semantic and structural features in the graph, respectively.
In particular, the semantic channel captures interaction between characters and words in the sentence without differentiating their types, and the structural channel adds the type and relative position when considering node interaction. The detailed definitions are given later. As shown in Figure \ref{fig:biaffine}, a graph neural network works by iteratively updating the representation of each node through layers. In each layer, each node receives information from its neighbors in the input graph structure so that its representation vector can be updated. The naive transformer structure \citep{vaswani2017attention} can be viewed as a graph attentional neural network, where each input character is a node and there is a graph edge between every two nodes.

Formally, denote the input of each layer as ${x}=[x_1, ..., x_n, {x}_{n+1}, ..., {x}_{n+k}]$, where $x_i$ is the input representation of $c_i$ for $1\leq i\leq n$ and  $x_i$ is the input representation of $w_{i-n}$ for $n+1\leq i\leq n+k$. 
The input for the first encoder layer is the embedding sequence 
$[\mathbf{e}'_{c_1}, ..., \mathbf{e}'_{c_n}, \mathbf{e}'_{w_{1}}, ..., \mathbf{e}'_{w_{k}}]$. 
For the $m$-th attention head, the similarity between two tokens  $\mathbf{x}_i$ and  $\mathbf{x}_j$ can be calculated by a vector inner product:
\begin{flalign*}
s^{sem,m}_{ji}=\left ( \mathbf{x}_i\mathbf{W}^{K,m} \right ) \cdot \left ( \mathbf{x}_j\mathbf{W}^{Q,m} \right ),
\end{flalign*}
where $\mathbf{W}^{K, m}$ and $\mathbf{W}^{Q,m}$ are the model parameters for the $m$-th attention head.

The similarity $s^{sem,m}_{ji}$ can control how much information $t_i$ can receive from $t_j$.
The key in the above information exchange process is a character-word mixed self-attention mechanism, where a weight score $s^{sem}_{ji}$  is calculated by the similarity of token $t_j$ and $t_i$. To this end, $s^{sem}_{ji}$ can be regarded as a \textbf{semantic channel}.

We further introduce a \textbf{structural channel} by taking  path structure into consideration. 
As shown in Table 1, we differentiate edges according to both the word/character difference and the relative position, resulting in 7 types of edges. For example, ``word $\rightarrow$ character'' represents an edge from a word to a character at its right. ``self-to-self'' represents a self-loop over a character or a word.

We make use of rich edge types by defining structural channels. 
Each token $t_i$ can receive the information from all the input according to the edges defined in Table \ref{tab:rels}.
Taking  the character ``华(China)'' as an example, it can receive information from ``成果(result)'', ``访(visit)'', ``成(become)'', ``华(China)'' through the edges $r_2$, $r_3$, $r_4$ and $r_5$ in the table, respectively. In addition, we define a special edge ``others'' for each other relation type so that ``华(China)'' can receive information from all characters and words. 

Formally, denote the relation from node $o$ to node $i$ as $r_{ji}$.
We can obtain the embedding of each relation $r_{ji}$ through an  embedding lookup table:
\begin{equation*}
\mathbf{e}_{ji} = \text{emb}(r_{ji}).
\end{equation*}

We further calculate contextualized relation embeddings $\mathbf{e}_{oi}^{ct}$, which is sensitive to the  representations of the two nodes $\mathbf{x}_i$ and $\mathbf{x}_o$ for $r_{oi}$:
\begin{flalign*}
\mathbf{e}^{ct, m}_{ji} = \sigma (\mathbf{e}_{{ji}}\mathbf{W}^{E,m} + \mathbf{e}_{x_i}\mathbf{W}^{S,m} + \mathbf{e}_{x_j}\mathbf{W}^{T,m}), 
\end{flalign*}
where $\mathbf{W}^{E,m}$, $\mathbf{W}^{S,m}$ and $\mathbf{W}^{T,m}$ are parameters. 

The structural similarity score is calculated as:
\begin{flalign*}
s^{str,m}_{ji} = \mathbf{w}^m \cdot \mathbf{e}^{ct,m}_{ji},
\end{flalign*}
where vector  $\mathbf{w}^m$ is a model parameter for the $m$-th attention head.

The final similarity score $s_{oi}$ between token $j \in [1, n+k] $ and token $i \in [1, n+k]$  is the combined scores from semantic channels and structural channels:
\begin{flalign*}
s_{ji}^m &= s^{sem,m}_{ji} + s^{str,m}_{ji}.
\end{flalign*}

The hidden state of $t_i$ for the $m$-th head is:
\begin{flalign*}
\mathbf{h}_i^m=\sum \textup{softmax}_j (\mathbf{s}_{ji}^{m}) \cdot \left ( \mathbf{x}_j\mathbf{W}^{V,m} \right ),
\end{flalign*}
where $\mathbf{x}_j$ is the input of token $t_j$ for the current encoder layer, and $\mathbf{W}^{V,m}$ is a model parameter.

\iftrue
The final output $\mathbf{h}_i$ is the concatenation of outputs from all the $M$ attention heads:
\begin{flalign*}
\mathbf{h}_i = [\mathbf{h}_i^1, ..., \mathbf{h}_i^M].
\end{flalign*}
\fi 

\section{Decoding and Training}

For word segmentation and POS tagging, we use a joint tag scheme $t_{ws}\_t_{pos}$, where $t_{ws} \in $ $ \left \{ B, M, E, S \right \}$ denotes segmentation labels \citep{xue2003chinese}, and $t_{pos}$ denotes POS. The probability $\textit{segpos}_{ij}$ for the joint tag 
$\left \{t_{ws}\_t_{pos} \right \}_i$ is:
\begin{flalign*}
\mathbf{t}_i=\text{MLP}_t(\mathbf{h}_{c_i});\ 
\textit{posseg}_{ij}=\text{softmax}_j\left ( \mathbf{W_t} \mathbf{t}_i \right )
\end{flalign*}

For dependency parsing, following \citet{dozat2016deep}, we use two representations of each character, which can distinguish between dependents and heads in dependency relations. In particular, for the $i$-th character $c_i$, a head representation $\mathbf{h}_{c_i}$ and a dependent representation $\mathbf{d}_{c_i}$ are obtained through two multi-layered perceptrons:
\begin{flalign*}
\mathbf{h}_i=\text{MLP}_h(\mathbf{h}_{c_i});\ \mathbf{d}_i=\text{MLP}_d(\mathbf{h}_{c_i})
\end{flalign*}

The dependency confidence score $dep_{ij}$ of the dependency relation $j\rightarrow i$ is obtained by using a biaffine transformation:
\begin{flalign*}
dep_{ij}=\text{softmax}_i\left ( \mathbf{h}_j^T\mathbf{A}\mathbf{d}_i + \mathbf{b}_1^T\mathbf{h}_j + \mathbf{b}_2^T\mathbf{d}_i \right )
\end{flalign*}
where $\mathbf{A}$, $\mathbf{b}_1$ and $\mathbf{b}_2$ are the model parameters.

\textbf{Training} A negative log-likelihood loss value $\mathcal{L}$ is computed on each character $c_i$ over the head probability $dep_{ij}$ locally and accumulated along the sentence,
\begin{equation*}
    \mathcal{L} = -\sum_{i=1}^{N} \left (  \log \textit{dep}_{i,head_i} + \log \textit{segpos}_{i,tag_{i}}\right),
\end{equation*}
where $head_i$ and $tag_i$ denotes the head index and the joint segmentation-POS tag of the $i$-th character, respectively. 
During training, we minimize $\mathcal{L}$ to train the model parameters. 

\textbf{Decoding} 
Hierarchical decoding is used for dependency parsing: 
we first perform decoding to parse the internal structure of a word and find the root character, and then perform decoding for all root character for all root character in the sentence to obtain a word-level parsing results.
Given the confidence score of all potential dependency arcs $dep_{ij}$, the decoding process can be formulated as a max spanning tree (MST) problem. Specifically, we use the Tarjan implementation \citep{tarjan1977finding} of the Chu-Liu-Edmonds algorithm \citep{chu1965shortest, edmonds1967optimum} to find the MST derivation.

\section{Experiments}
We investigate the effect of word information and our GAT encoder to character-level graph-based Chinese parsing.

\subsection{Settings}

We use three releases of the Chinese Penn Treebank (i.e., 5.0, 6.0 and 7.0) \citep{xue2005penn},
splitting the corpora into training, development and test sets according to previous work. 
CTB 5.0 is split by following \citet{zhang-clark:2010:EMNLP}, CTB 6.0 by following the official documentation, and CTB 7.0 by following \citet{wang2011improving}. 

We use the head rules of \citet{zhang-clark:2008:EMNLP} to convert phrase structures in CTB into dependency structures. 
Following \citet{hatori2012incremental}, the standard measures of word-level precision, recall and F1 score are used to evaluate word segmentation, POS-tagging and dependency parsing, respectively. 
For a given word $w=c_1c_2...c_k$, we simply use the right branching tree $c_1 \rightarrow c_2 \rightarrow ... \rightarrow c_k$ as the intra-word dependency structure.
Following \citet{zhang-yang-2018-chinese}, we take the external Chinese lexicon dictionary from \citet{song2018directional}.
The experiments are conducted under the same hardware environment: Geforce GTX 2080Ti graph card and i7-7900 CPU. 

\textbf{Hyper-parameters} We initialize character embeddings using BERT \citep{devlin2018bert}, and 
word embeddings with the average pooling result of BERT character embeddings.
Neither character nor word embeddings are fine-tuned due to limitation of GPU memory. The size of hidden states is set to 400 for all models (The hidden size of each direction in biLSTM is 200).

Training is done on mini-baches via Adagrad \citep{duchi2011adaptive} with a learning rate of 0.002, $\beta_1=0.9, \beta_2=0.9$ and $\epsilon=1e-12$. 
We adopt gradient clipping with a threshold of 5.0. Dropout \citep{hinton2012improving} is used on each layer, with a rate of 0.2.

\subsection{Development Experiments}

We conduct development experiments on the CTB 5.0 dev set to decide the final hyper-parameters of the two mixed model. Below we show the details of three important factors.

\textbf{Effect of Encoder Layers} 
We investigate the effect of the number of encoder layers for the word-character graph attention model. The results are shown in Table \ref{tab:dev_lattice_att}. As the number of encoder layer increases from 1 to 3, performance improves for parsing, as well as word segmentation and POS tagging. We do not conduct experiments on more layers because of memory limitation. The number of encoder layers is set to 3 in the remaining experiments.

\textbf{Effect of Attention Heads} We further investigate the effect of attention head number for the word-character graph attention  model, as shown in Table \ref{tab:dev_lattice_att}.  We observe performance increases as the number of attention heads increases from 1
to 4, but adding more attention heads does not lead a further performance improvement. We thus fix the number of attention heads to 4 accordingly.

\begin{table}[t]
\small
\centering
\begin{tabular}{llll}
\ttrule
\textbf{Parameter} & \textbf{SEG}  & \textbf{POS} & \textbf{DEP}    \\ 
\midrule
1 head & 98.5 & 96.4 & 91.0 \\
4 heads & 98.8 & 96.5 & 91.4 \\
8 heads & 98.6 & 96.4 & 91.3 \\
16 heads & 98.7 & 96.5 & 90.9 \\ \hline
1 layer & 98.5 & 96.2 & 90.0 \\
2 layers & 98.5 & 96.3 & 91.1  \\
3 layers & 98.8 & 96.5 & 91.4  \\ 
\ddrule
\end{tabular}
\caption{Dev experiments for word-character graph attention model.}
\label{tab:dev_lattice_att}
\end{table}

\textbf{Effect of Word Information and Model Architectures} 
We conduct a set of development experiments to verify the influence of model architecture and word information on each architecture.

The results are shown in Table \ref{tab:dev}. In particular, a character-level LSTM model gives a parsing accuracy of 90.3\%. With word-character lattice, the results are improved to 90.8\%
Using GAT, the model gives a 90.9\% development accuracy for parsing, which is improved to 91.4\% by using word information additionally. 
This shows that word information is beneficial to both the LSTM architecture and the GAT architecture. In addition to dependency structures, the results for both word segmentation and POS-tagging are also improved. 

In addition, LSTM models underperform their SAN counterparts regardless whether word information is combined into the character encoder. 
The lattice LSTM system is much slower to train (nearly half an hour for training one epoch on CTB 5, and the total training takes several days) compared with the chracter-level BiLSTM model (145 seconds for one epoch) and  SAN models (421 seconds for one epoch).

\begin{table}[t]
\small
\centering
\begin{tabular}{llll}
\toprule
\textbf{Model} & \textbf{SEG}  & \textbf{POS} & \textbf{DEP}    \\ 
\midrule
Character-level LSTM &  98.4 & 96.1 & 90.3 \\
Word-char lattice LSTM  &  98.7 &	96.4  & 90.8 \\
Character-level transformer & 98.7 &	96.5  & 90.9    \\
Word-char graph attention & 98.8 &	96.5 & 91.4  \\ 
\ddrule
\end{tabular}
\caption{Dev experiments on different models.}
\label{tab:dev}
\end{table}

\begin{table*}[tp]
\small
  \centering
    \begin{tabular}{lccccccccc}
    \specialrule{.05em}{0pt} {.65ex}
    \multirow{2}{*}{Model}&
    \multicolumn{3}{c}{CTB 5.0}&\multicolumn{3}{c}{CTB 6.0}&\multicolumn{3}{c}{CTB 7.0}\cr
    \cmidrule(lr){2-4} \cmidrule(lr){5-7} \cmidrule(lr){8-10}
    &SEG & POS & DEP &SEG & POS & DEP & SEG & POS & DEP \cr
    \midrule
    Incremental Joint \citep{hatori2012incremental} & 96.9  & 93.0 & 76.0  & 96.2  & 92.0  & 75.8 & 96.1  & 91.3 & 74.6 \\
    Char STD \citep{zhang2014character} & 97.8 & 94.6 & 82.1  & 95.6   & 91.4 & 77.1 & 95.5 & 90.8 & 75.7 \\
    Char EAG \citet{zhang2014character} & 97.8 & 94.4 & 82.1  & 95.7 & 91.5 & 77.0 & 95.5 & 90.7 & 75.8 \\
    Joint Annotated \citep{zhang2015randomized} & 98.0 & 94.5 & 82.0 & - & - & - & - & - & - \\
    NN transition \citep{kurita2017neural} & 98.4 & 94.8 & 81.4 & - & - & - & 96.4 & 91.3 & 75.3 \\ 
    NN char-level \citep{li2018neural} & 96.6 & 92.9 & 79.4* & - & - & - & - & - & - \\
    Joint Multi BERT \citep{yan2019unified} & 98.5 & - & 89.6 & - &- & - & 97.1 & - & 85.1 \\
    \midrule
    Lattice-LSTM E2E BERT & 98.4 & 96.2 & 88.1 & 97.3 & 94.6 & 83.6 & 96.9 & 93.9 & 84.9 \\
    \textbf{Word-char Graph Attention} & \textbf{98.7} & \textbf{96.5} & \textbf{91.3} & \textbf{97.3} & \textbf{94.8} & \textbf{87.2} & \textbf{97.3} & \textbf{94.4} & \textbf{86.2} \\
    \specialrule{.05em}{.4ex}{0pt}
    \end{tabular}
\caption{Final Results. * indicate performance on CTB 5.1.}
\label{tab:result}
\end{table*}

\subsection{Final Results}
Table \ref{tab:result} shows the overall performances of end-to-end Chinese dependency parsing where our model is compared with the state-of-the-art methods in the literature.
We report the performances on Chinese word segmentation, POS tagging and parsing, respectively. 
First, we can find that BERT pretrained models outperform neural models using traditional embeddings \citep{kurita2017neural, li2018neural} and statistical models  \citep{hatori2012incremental, zhang2014character, zhang2015randomized} by a large margin. 
Compared with Joint Multi BERT \citep{yan2019unified}, the word-char graph attention model achieves better performance especially on parsing.
Note that we do not leverage on additional syntactic and lexical annotation within a word as \citet{zhang2014character} and \citet{li2018neural} do. 
This demonstrates the advantage of our word-character graph attention model.

The final results on parsing of the lattice-LSTM BERT end-to-end model are 88.1\%, 83.6\% and 84.9\% on the CTB 5, 6 and 7 datasets, respectively, and the final results of the GAT model are 91.3\%, 87.2\% and 86.2\% on the three datasets, respectively. The improvements are statistically significant at $p < 0.05$ using t-test. Overall, the GAT architecture gives better segmentation, POS-tagging and parsing results compared with the LSTM model on all datasets, while running much faster. This shows the advantage of our semantic and structural channels for integrating word information. Compared with the existing methods for joint Chinese parsing, our final GAT model gives better results on all the three datasets, achieving the best reported accuracies on dependency parsing in the literature.

\textbf{Segmentation Results} 
According to Table~\ref{tab:result}, word information brings the most improvements on parsing accuracies, followed by tagging accuracies. Segmentation benefits relatively less.
However, as a fully end-to-end model, POS tagging and dependency parsing information can also benefit word segmentation. Thus our joint model can be a competitive choice for the segmentation task alone. 
We compare the performance of different models with BERT pretraining on the CTB6 word segmentation task in Table \ref{tab:exp:cws_test} (previous work mostly uses the CTB6 version). 

The first 4 items are the segmentation models trained on the single segmentation task.  BiLSTM-CRF-BERT \citep{gan2019investigating} uses a BiLSTM network followed by CRF network with pretrained BERT as input. LSAN-CRF \citep{gan2019investigating} uses a local self-attention network instead of BiLSTM. DP-BERT \citep{huang2019toward} and Unified-BERT \citep{ke2020unified} are both multi-criteria Chinese word segmentation models, which are trained on a range of extra training sets to enhance the segmentor. DP-BERT \citep{huang2019toward} uses a domain projection for multi-criteria learning while unified BERT \citep{ke2020unified} 
employs a fully shared model for all criteria. 

Our model gives better performance on segmentation BiLSTM-CRF-BERT and a bit worse result than LSAN-CRF-BERT, which is optimised for segmentation.
In addition, our model gives comparable results in segmentation compared with DP-BERT and Unified-BERT, 
which shows the advantage of joint parsing as compared to external segmentation datasets. 
Compared with \citet{yan2019unified}, our model can give better segmentation.

\subsection{Analysis}
\label{sec:analysis}

In this section, we focus on the word-character GAT model to thoroughly investigate the effect of word information.

\begin{table}[t]
\small
\centering
\begin{tabular}{lc} 
\ttrule
Model & F1-score \\  \midrule
BiLSTM-CRF-BERT \citep{gan2019investigating}  & 97.2  \\
LSAN-CRF-BERT \citep{gan2019investigating} & 97.4\\
DP-BERT \citep{huang2019toward} & 97.6 \\
Unified BERT \citep{ke2020unified}  & 97.2\\ %
Word-char Graph Attention  & 97.3  \\ 
\ddrule
\end{tabular}
\caption{Word Segmentation results. All results are based on BERT pretraining.}
\label{tab:exp:cws_test}
\end{table}

\begin{figure}[!tb]
   \centering
   \includegraphics[width=0.85\linewidth]{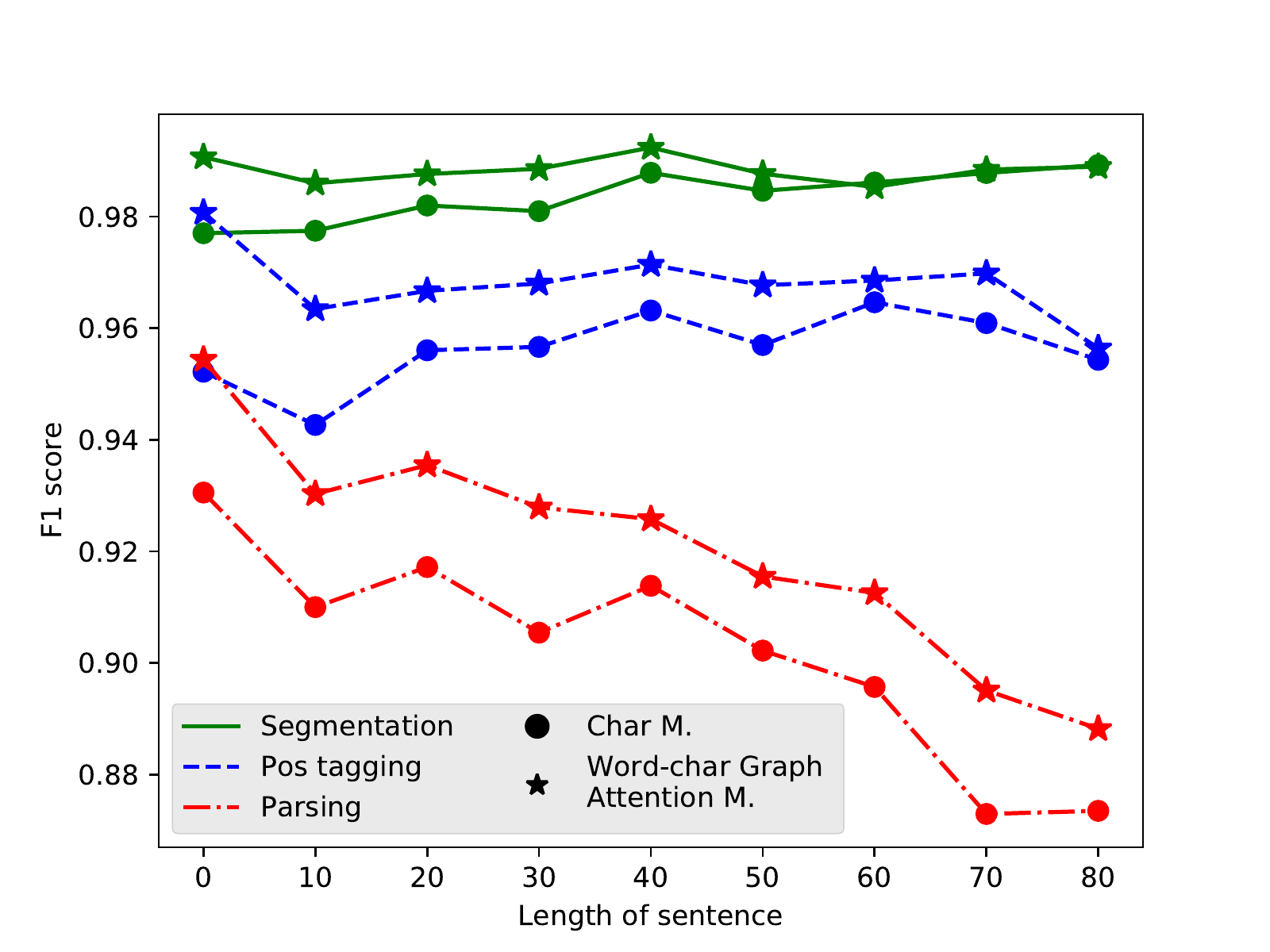}
   \caption{Performance against sentence length. The F1 score for each length $l$ is calculated on the test set sentences length in the bin $[l, l+15]$.}
      \label{fig:exp:lens}
\end{figure}

\begin{figure*}[!tb]  
 \setlength{\abovecaptionskip}{4pt}
  \centering
  \includegraphics[width=0.96\linewidth]{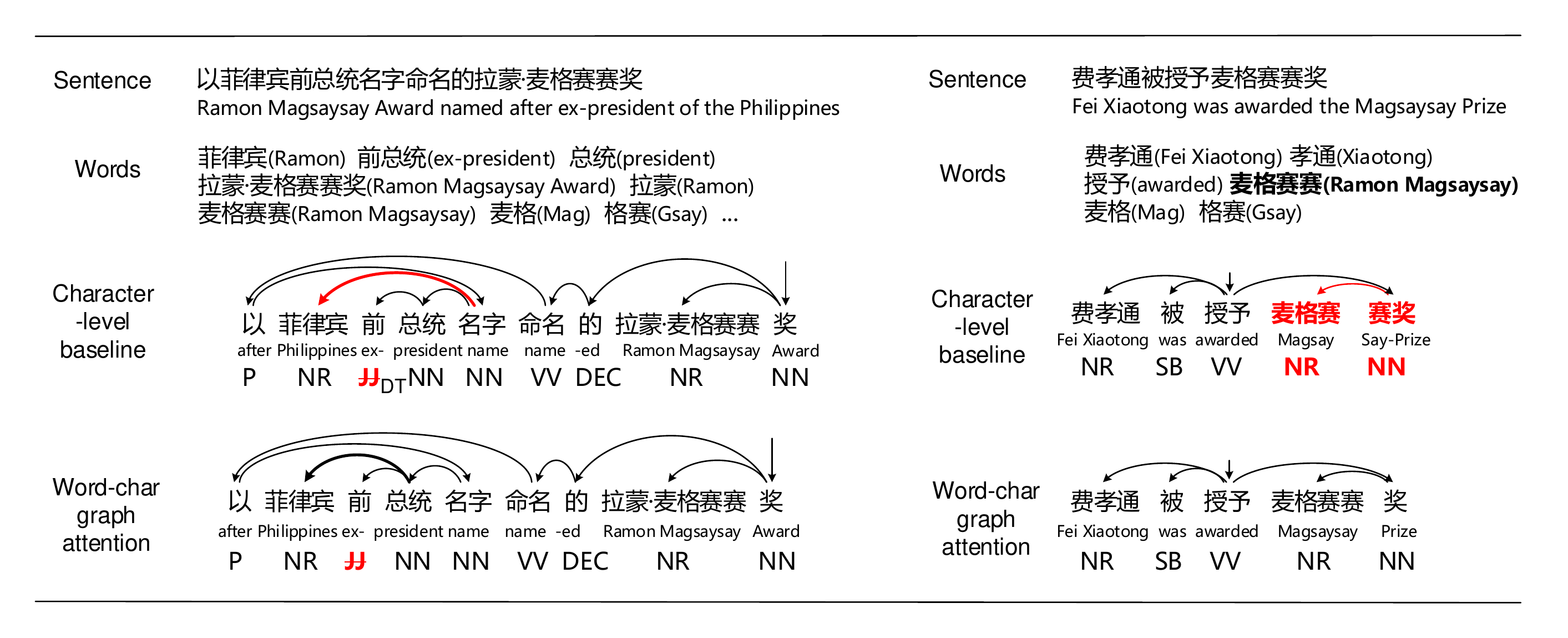}
  \caption{Sample outputs. Errors are in red.}
      \label{fig:exp:cases}
\end{figure*}

\textbf{OOV}
Table \ref{tab:exp:oov} shows the recall of out-of-vocabulary words on the CTB 5.0 test set. A word-character graph attention model enriched with word information achieves 0.9\%, 2.1\% and 4.1\% absolute improvement over the character-level model on word segmentation, POS tagging and parsing, respectively. It shows that the word-character attention model benefits transfer learning and generalization. 
This is likely because lexicons help to learn better representation for OOV words.
In particular, the word-char attention model shows significant improvement on parsing. The improvement of parsing results from better segmentation (+ 0.9\%) and better head inference (+3.8\%).

\begin{table}[t]
\small
\centering
\begin{tabular}{lccc}
\ttrule
Task & Char M. & Word-char M. & Diff. \\  
\midrule %
SEG & 87.4 & 88.3 & +0.9 \\
POS & 81.7 & 83.8 & +2.1 \\
DEP & 76.6 & 80.7 & +4.1 \\
DEP$'$ & 87.6 & 91.4 & +3.8 \\
\ddrule %
\end{tabular}
\caption{Recalls of OOV words. DEP$'$ indicates the parsing recall rate when the dependent word is correctly segmented.  }
\label{tab:exp:oov}
\end{table}

\textbf{Performance Against the Sentence Length}
We further analyze the performance with respect to different sentence lengths. Figure \ref{fig:exp:lens} shows the results. The performance of word segmentation and POS tagging are relatively stable against sentence length as these two tasks rely more on the local context. There is a tendency of decreasing F1 scores for parsing when the sentence length increases.
Compared with the the Transformer model without using word information, the word-character GAT model gives about the same improvements on parsing across different sentence lengths. In addition, the use of word information also makes the curve more smooth across different sentence lengths, which shows the benefit of lexicon features in enhancing the model robustness.

\textbf{Ablation Study}
We conduct ablation experiments to investigate the effect of different channels  of the word-char graph attention model showing three important factors on the CTB 5.0 test set in Table \ref{tab:exp:ablation}. 
Compared with the word-char graph attention model, removing the structural channel leads to a  parsing F1 score decrease by 0.6\%. In contrast, excluding the semantic channel decreases the parsing F1 score by 0.8\%.
Compared with the structural channel, the semantic channel contains important position information besides semantic similarity, which can be the reason for its better performance. However, 
the best result is achieved using both channels, from which it can be inferred that the channels can  interact with and complement each other on the final performance.
We can also observe that using word information through either channel or both channels combined can improve the parsing performance compared to a character-level attention model (``- all word info'' in the table), showing the effectiveness of word information.

\begin{table}[t]
\small
\centering
\begin{tabular}{lcc}
\hline
\textbf{Configuration} & \textbf{F1 score} & \textbf{Diff.} \\  
\hline 
Word-char graph attention & 91.3 & 0 \\
- structural channel & 90.8 & -0.5 \\
- semantic channel & 90.6 & -0.7 \\
- all word information & 90.5 & -0.8 \\
\hline
\end{tabular}
\caption{Ablation test of the  word-character graph attention model.}
\label{tab:exp:ablation}
\end{table}

\textbf{Case Study}
Figure \ref{fig:exp:cases} shows samples of end-to-end Chinese parsing on CTB 5.0. In the first sentence, the head word of token ``菲律宾(the Philippines)'' is incorrectly recognized as ``名字(name)'' in the vanilla LSTM model, while the word-char graph attention model identifies its correct head ``总统(president)''. 
The vanilla LSTM model suffers from polysemous characters
``菲(commonly used character in names)'', ``律(law)'' and ``宾(commonly used character in names)'' . 
In contrast, the word-character graph attention model taking additional word information ``菲律宾(the Philippines)'' can better understand the sentence and thus gives a correct syntactic parsing result.

In the second sentence, the expression ``麦格赛赛奖(Magsaysay Award)'' (Gold segmentation: ``麦格赛赛(Magsaysay)'', ``奖(Award)'') is incorrectly segmented into tokens ``麦格赛(Magsay)'' and ``赛奖(Competition Award)'' by the vanilla LSTM model, which results in errors in the POS tagging and parsing result. It shows that word features can help better identify word boundaries, which is important
to POS tagging and parsing. The word-char graph attention model segments the sentence correctly, and thus gives the correct result in all three tasks. In summary, the word-character graph attention model  performs better in identifying dependency heads in addition to word boundaries, thanks to the lexicon input.

\vspace{-1mm}
\section{Conclusion}
\label{sec:con}
We investigated the effectiveness of word information for end-to-end graph-based Chinese parsing by 
extending a character-level transformer encoder. Compared with LSTM-based representation learning models, our method is more feasible due to better parallelization between characters and convenience to batching.
Results on the three datasets show that word information benefits all the three tasks, and 
our word-char graph attention model outperforms the lattice-LSTM model, 
and achieving the best results on segmentation, POS tagging and parsing in the literature. 
To our knowledge, we are the first to investigate an end-to-end Chinese parser exploiting lexicon knowledge.

\bibliography{ref}

\begin{thebibliography}{53}
\expandafter\ifx\csname natexlab\endcsname\relax\def\natexlab#1{#1}\fi

\bibitem[{Andor et~al.(2016)Andor, Alberti, Weiss, Severyn, Presta, Ganchev,
  Petrov, and Collins}]{andor:2016:global}
Daniel Andor, Chris Alberti, David Weiss, Aliaksei Severyn, Alessandro Presta,
  Kuzman Ganchev, Slav Petrov, and Michael Collins. 2016.
\newblock \href {http://www.aclweb.org/anthology/P16-1231} {Globally normalized
  transition-based neural networks}.
\newblock In \emph{Proceedings of the 54th Annual Meeting of the Association
  for Computational Linguistics (Volume 1: Long Papers)}, pages 2442--2452,
  Berlin, Germany. Association for Computational Linguistics.

\bibitem[{Bohnet(2010)}]{bohnet:2010:coling}
Bernd Bohnet. 2010.
\newblock \href {http://www.aclweb.org/anthology/C10-1011} {Top accuracy and
  fast dependency parsing is not a contradiction}.
\newblock In \emph{Proceedings of the 23rd International Conference on
  Computational Linguistics (Coling 2010)}, pages 89--97, Beijing, China.
  Coling 2010 Organizing Committee.

\bibitem[{Chen and Manning(2014)}]{chen2014dep}
Danqi Chen and Christopher Manning. 2014.
\newblock A fast and accurate dependency parser using neural networks.
\newblock In \emph{Proceedings of the 2014 Conference on Empirical Methods in
  Natural Language Processing (EMNLP)}.

\bibitem[{Chen et~al.(2017)Chen, Shi, Qiu, and Huang}]{chen2017dag}
Xinchi Chen, Zhan Shi, Xipeng Qiu, and Xuanjing Huang. 2017.
\newblock Dag-based long short-term memory for neural word segmentation.
\newblock \emph{arXiv preprint arXiv:1707.00248}.

\bibitem[{Chu(1965)}]{chu1965shortest}
Yoeng-Jin Chu. 1965.
\newblock On the shortest arborescence of a directed graph.
\newblock \emph{Scientia Sinica}.

\bibitem[{Devlin et~al.(2018)Devlin, Chang, Lee, and
  Toutanova}]{devlin2018bert}
Jacob Devlin, Ming-Wei Chang, Kenton Lee, and Kristina Toutanova. 2018.
\newblock Bert: Pre-training of deep bidirectional transformers for language
  understanding.
\newblock \emph{arXiv preprint arXiv:1810.04805}.

\bibitem[{Ding et~al.(2019)Ding, Xie, Zhang, Lu, Li, and
  Si}]{ding-etal-2019-neural}
Ruixue Ding, Pengjun Xie, Xiaoyan Zhang, Wei Lu, Linlin Li, and Luo Si. 2019.
\newblock A neural multi-digraph model for {C}hinese {NER} with gazetteers.
\newblock In \emph{Proceedings of the 57th Annual Meeting of the Association
  for Computational Linguistics}.

\bibitem[{Dozat and Manning(2016)}]{dozat2016deep}
Timothy Dozat and Christopher~D Manning. 2016.
\newblock Deep biaffine attention for neural dependency parsing.
\newblock \emph{arXiv preprint arXiv:1611.01734}.

\bibitem[{Dozat et~al.(2017)Dozat, Qi, and Manning}]{dozat-etal-2017-stanfords}
Timothy Dozat, Peng Qi, and Christopher~D. Manning. 2017.
\newblock {S}tanford{'}s graph-based neural dependency parser at the {C}o{NLL}
  2017 shared task.
\newblock In \emph{Proceedings of the {C}o{NLL}}.

\bibitem[{Duchi et~al.(2011)Duchi, Hazan, and Singer}]{duchi2011adaptive}
John Duchi, Elad Hazan, and Yoram Singer. 2011.
\newblock Adaptive subgradient methods for online learning and stochastic
  optimization.
\newblock \emph{Journal of Machine Learning Research}.

\bibitem[{Edmonds(1967)}]{edmonds1967optimum}
Jack Edmonds. 1967.
\newblock Optimum branchings.
\newblock \emph{Journal of Research of the national Bureau of Standards B},
  71(4):233--240.

\bibitem[{Gamallo et~al.(2012)Gamallo, Garcia, and
  Fern{\'a}ndez-Lanza}]{gamallo2012dependency}
Pablo Gamallo, Marcos Garcia, and Santiago Fern{\'a}ndez-Lanza. 2012.
\newblock Dependency-based open information extraction.
\newblock In \emph{Proceedings of the joint workshop on unsupervised and
  semi-supervised learning in NLP}.

\bibitem[{Gan and Zhang(2019)}]{gan2019investigating}
Leilei Gan and Yue Zhang. 2019.
\newblock Investigating self-attention network for chinese word segmentation.
\newblock \emph{arXiv preprint arXiv:1907.11512}.

\bibitem[{Guo et~al.(2019)Guo, Zhang, and Lu}]{guo-etal-2019-attention}
Zhijiang Guo, Yan Zhang, and Wei Lu. 2019.
\newblock Attention guided graph convolutional networks for relation
  extraction.
\newblock In \emph{Proceedings of ACL}.

\bibitem[{Haji{\v{c}} et~al.(2009)Haji{\v{c}}, Ciaramita, Johansson, Kawahara,
  Mart{\'\i}, M{\`a}rquez, Meyers, Nivre, Pad{\'o}, {\v{S}}t{\v{e}}p{\'a}nek,
  Stra{\v{n}}{\'a}k, Surdeanu, Xue, and Zhang}]{hajic-etal-2009-conll}
Jan Haji{\v{c}}, Massimiliano Ciaramita, Richard Johansson, Daisuke Kawahara,
  Maria~Ant{\`o}nia Mart{\'\i}, Llu{\'\i}s M{\`a}rquez, Adam Meyers, Joakim
  Nivre, Sebastian Pad{\'o}, Jan {\v{S}}t{\v{e}}p{\'a}nek, Pavel
  Stra{\v{n}}{\'a}k, Mihai Surdeanu, Nianwen Xue, and Yi~Zhang. 2009.
\newblock The {C}o{NLL}-2009 shared task: Syntactic and semantic dependencies
  in multiple languages.
\newblock In \emph{Proceedings of {C}o{NLL}}.

\bibitem[{Hatori et~al.(2012)Hatori, Matsuzaki, Miyao, and
  Tsujii}]{hatori2012incremental}
Jun Hatori, Takuya Matsuzaki, Yusuke Miyao, and Jun'ichi Tsujii. 2012.
\newblock Incremental joint approach to word segmentation, pos tagging, and
  dependency parsing in chinese.
\newblock In \emph{Proceedings of ACL}.

\bibitem[{Hinton et~al.(2012)Hinton, Srivastava, Krizhevsky, Sutskever, and
  Salakhutdinov}]{hinton2012improving}
Geoffrey~E Hinton, Nitish Srivastava, Alex Krizhevsky, Ilya Sutskever, and
  Ruslan~R Salakhutdinov. 2012.
\newblock Improving neural networks by preventing co-adaptation of feature
  detectors.
\newblock \emph{arXiv preprint arXiv:1207.0580}.

\bibitem[{Huang et~al.(2019)Huang, Cheng, Chen, Wang, and
  Chu}]{huang2019toward}
Weipeng Huang, Xingyi Cheng, Kunlong Chen, Taifeng Wang, and Wei Chu. 2019.
\newblock Toward fast and accurate neural chinese word segmentation with
  multi-criteria learning.
\newblock \emph{arXiv preprint arXiv:1903.04190}.

\bibitem[{Ke et~al.(2020)Ke, Shi, Meng, Wang, Qiu, and Huang}]{ke2020unified}
Zhen Ke, Liang Shi, Erli Meng, Bin Wang, Xipeng Qiu, and Xuanjing Huang. 2020.
\newblock Unified multi-criteria chinese word segmentation with bert.
\newblock \emph{arXiv preprint arXiv:2004.05808}.

\bibitem[{Kiperwasser and Goldberg(2016)}]{kiperwasser2016simple}
Eliyahu Kiperwasser and Yoav Goldberg. 2016.
\newblock Simple and accurate dependency parsing using bidirectional lstm
  feature representations.
\newblock \emph{Transactions of the Association for Computational Linguistics}.

\bibitem[{Kipf and Welling(2016)}]{kipf2016semi}
Thomas~N Kipf and Max Welling. 2016.
\newblock Semi-supervised classification with graph convolutional networks.
\newblock \emph{arXiv preprint arXiv:1609.02907}.

\bibitem[{Kurita et~al.(2017)Kurita, Kawahara, and
  Kurohashi}]{kurita2017neural}
Shuhei Kurita, Daisuke Kawahara, and Sadao Kurohashi. 2017.
\newblock Neural joint model for transition-based chinese syntactic analysis.
\newblock In \emph{Proceedings of ACL}, pages 1204--1214.

\bibitem[{Li et~al.(2018)Li, Zhang, Ju, and Zhao}]{li2018neural}
Haonan Li, Zhisong Zhang, Yuqi Ju, and Hai Zhao. 2018.
\newblock Neural character-level dependency parsing for chinese.
\newblock In \emph{Thirty-Second AAAI Conference on Artificial Intelligence}.

\bibitem[{Li et~al.(2019)Li, Li, Zhang, Wang, Li, and Si}]{li2019self}
Ying Li, Zhenghua Li, Min Zhang, Rui Wang, Sheng Li, and Luo Si. 2019.
\newblock Self-attentive biaffine dependency parsing.
\newblock In \emph{Proceedings of IJCAI}.

\bibitem[{Ma et~al.(2018)Ma, Hu, Liu, Peng, Neubig, and Hovy}]{ma2018stack}
Xuezhe Ma, Zecong Hu, Jingzhou Liu, Nanyun Peng, Graham Neubig, and Eduard
  Hovy. 2018.
\newblock Stack-pointer networks for dependency parsing.
\newblock \emph{arXiv preprint arXiv:1805.01087}.

\bibitem[{Ma and Zhao(2012)}]{ma2012fourth}
Xuezhe Ma and Hai Zhao. 2012.
\newblock Fourth-order dependency parsing.
\newblock In \emph{Proceedings of COLING 2012: posters}, pages 785--796.

\bibitem[{Miwa and Bansal(2016)}]{miwa-bansal-2016-end}
Makoto Miwa and Mohit Bansal. 2016.
\newblock End-to-end relation extraction using {LSTM}s on sequences and tree
  structures.
\newblock In \emph{Proceedings of ACL}.

\bibitem[{Nivre(2008)}]{Nivre:2008:CL}
Joakim Nivre. 2008.
\newblock \href {https://doi.org/10.1162/coli.07-056-R1-07-027} {Algorithms for
  deterministic incremental dependency parsing}.
\newblock \emph{Comput. Linguist.}, 34(4):513--553.

\bibitem[{Poon and Domingos(2009)}]{poon-domingos-2009-unsupervised}
Hoifung Poon and Pedro Domingos. 2009.
\newblock Unsupervised semantic parsing.
\newblock In \emph{Proceedings of the 2009 Conference on Empirical Methods in
  Natural Language Processing}.

\bibitem[{Song et~al.(2018)Song, Shi, Li, and Zhang}]{song2018directional}
Yan Song, Shuming Shi, Jing Li, and Haisong Zhang. 2018.
\newblock Directional skip-gram: Explicitly distinguishing left and right
  context for word embeddings.
\newblock In \emph{Proceedings of the 2018 Conference of the North American
  Chapter of the Association for Computational Linguistics: Human Language
  Technologies, Volume 2 (Short Papers)}, pages 175--180.

\bibitem[{Sperber et~al.(2019)Sperber, Neubig, Pham, and
  Waibel}]{sperber-etal-2019-self}
Matthias Sperber, Graham Neubig, Ngoc-Quan Pham, and Alex Waibel. 2019.
\newblock Self-attentional models for lattice inputs.
\newblock In \emph{Proceedings of the 57th Annual Meeting of the Association
  for Computational Linguistics}.

\bibitem[{Strubell et~al.(2018)Strubell, Verga, Andor, Weiss, and
  McCallum}]{strubell-etal-2018-linguistically}
Emma Strubell, Patrick Verga, Daniel Andor, David Weiss, and Andrew McCallum.
  2018.
\newblock Linguistically-informed self-attention for semantic role labeling.
\newblock In \emph{Proceedings of EMNLP}.

\bibitem[{Sun et~al.(2018)Sun, Tang, Duan, Ji, Cao, Feng, Qin, Liu, and
  Zhou}]{sun-etal-2018-semantic}
Yibo Sun, Duyu Tang, Nan Duan, Jianshu Ji, Guihong Cao, Xiaocheng Feng, Bing
  Qin, Ting Liu, and Ming Zhou. 2018.
\newblock Semantic parsing with syntax- and table-aware {SQL} generation.
\newblock In \emph{Proceedings of the 56th Annual Meeting of the Association
  for Computational Linguistics (Volume 1: Long Papers)}.

\bibitem[{Tarjan(1977)}]{tarjan1977finding}
Robert~Endre Tarjan. 1977.
\newblock Finding optimum branchings.
\newblock \emph{Networks}.

\bibitem[{Vaswani et~al.(2017)Vaswani, Shazeer, Parmar, Uszkoreit, Jones,
  Gomez, Kaiser, and Polosukhin}]{vaswani2017attention}
Ashish Vaswani, Noam Shazeer, Niki Parmar, Jakob Uszkoreit, Llion Jones,
  Aidan~N Gomez, {\L}ukasz Kaiser, and Illia Polosukhin. 2017.
\newblock Attention is all you need.
\newblock In \emph{NIPS}.

\bibitem[{Veličković et~al.(2017)Veličković, Cucurull, Casanova, Romero,
  Liò, and Bengio}]{velikovi2017graph}
Petar Veličković, Guillem Cucurull, Arantxa Casanova, Adriana Romero, Pietro
  Liò, and Yoshua Bengio. 2017.
\newblock \href {http://arxiv.org/abs/1710.10903} {Graph attention networks}.

\bibitem[{Wang and Chang(2016)}]{wang-chang-2016-graph}
Wenhui Wang and Baobao Chang. 2016.
\newblock Graph-based dependency parsing with bidirectional {LSTM}.
\newblock In \emph{Proceedings of ACL}.

\bibitem[{Wang et~al.(2011)Wang, Kazama, Tsuruoka, Chen, Zhang, and
  Torisawa}]{wang2011improving}
Yiou Wang, Jun’ichi Kazama, Yoshimasa Tsuruoka, Wenliang Chen, Yujie Zhang,
  and Kentaro Torisawa. 2011.
\newblock Improving chinese word segmentation and pos tagging with
  semi-supervised methods using large auto-analyzed data.
\newblock In \emph{Proceedings of 5th International Joint Conference on Natural
  Language Processing}, pages 309--317.

\bibitem[{Xue et~al.(2005)Xue, Xia, Chiou, and Palmer}]{xue2005penn}
Naiwen Xue, Fei Xia, Fu-Dong Chiou, and Marta Palmer. 2005.
\newblock The penn chinese treebank: Phrase structure annotation of a large
  corpus.
\newblock \emph{Natural language engineering}, 11(02):207--238.

\bibitem[{Xue and Shen(2003)}]{xue2003chinese}
Nianwen Xue and Libin Shen. 2003.
\newblock Chinese word segmentation as lmr tagging.
\newblock In \emph{Proceedings of the second SIGHAN workshop on Chinese
  language processing-Volume 17}, pages 176--179. Association for Computational
  Linguistics.

\bibitem[{Yan et~al.(2020)Yan, Qiu, and Huang}]{yan2019unified}
Hang Yan, Xipeng Qiu, and Xuanjing Huang. 2020.
\newblock A graph-based model for joint chinese word segmentation and
  dependency parsing.
\newblock \emph{Transactions of the Association for Computational Linguistics},
  8:78--92.

\bibitem[{Zhang and McDonald(2014)}]{zhangmcdonald2014enforcing}
Hao Zhang and Ryan McDonald. 2014.
\newblock Enforcing structural diversity in cube-pruned dependency parsing.
\newblock In \emph{Proceedings of the 52nd Annual Meeting of the Association
  for Computational Linguistics (Volume 2: Short Papers)}.

\bibitem[{Zhang et~al.(2013)Zhang, Zhang, Che, and Liu}]{zhang2013chinese}
Meishan Zhang, Yue Zhang, Wanxiang Che, and Ting Liu. 2013.
\newblock Chinese parsing exploiting characters.
\newblock In \emph{Proceedings of ACL}.

\bibitem[{Zhang et~al.(2014)Zhang, Zhang, Che, and Liu}]{zhang2014character}
Meishan Zhang, Yue Zhang, Wanxiang Che, and Ting Liu. 2014.
\newblock Character-level chinese dependency parsing.
\newblock In \emph{Proceedings of ACL}.

\bibitem[{Zhang et~al.(2019)Zhang, Ge, Chen, and Fan}]{zhang-etal-2019-lattice}
Pei Zhang, Niyu Ge, Boxing Chen, and Kai Fan. 2019.
\newblock Lattice transformer for speech translation.
\newblock In \emph{Proceedings of the 57th Annual Meeting of the Association
  for Computational Linguistics}.

\bibitem[{Zhang et~al.(2015)Zhang, Li, Barzilay, and
  Darwish}]{zhang2015randomized}
Yuan Zhang, Chengtao Li, Regina Barzilay, and Kareem Darwish. 2015.
\newblock Randomized greedy inference for joint segmentation, pos tagging and
  dependency parsing.

\bibitem[{Zhang and Clark(2008)}]{zhang-clark:2008:EMNLP}
Yue Zhang and Stephen Clark. 2008.
\newblock A tale of two parsers: {I}nvestigating and combining graph-based and
  transition-based dependency parsing.
\newblock In \emph{Proceedings of EMNLP}.

\bibitem[{Zhang and Clark(2010)}]{zhang-clark:2010:EMNLP}
Yue Zhang and Stephen Clark. 2010.
\newblock A fast decoder for joint word segmentation and {POS}-tagging using a
  single discriminative model.
\newblock In \emph{Proceedings of the 2010 Conference on Empirical Methods in
  Natural Language Processing}.

\bibitem[{Zhang and Nivre(2011)}]{zhangnivre2011transition}
Yue Zhang and Joakim Nivre. 2011.
\newblock \href {http://www.aclweb.org/anthology/P11-2033} {Transition-based
  dependency parsing with rich non-local features}.
\newblock In \emph{Proceedings of the 49th Annual Meeting of the Association
  for Computational Linguistics: Human Language Technologies}, pages 188--193,
  Portland, Oregon, USA. Association for Computational Linguistics.

\bibitem[{Zhang and Yang(2018)}]{zhang-yang-2018-chinese}
Yue Zhang and Jie Yang. 2018.
\newblock {C}hinese {NER} using lattice {LSTM}.
\newblock In \emph{Proceedings of ACL}.

\bibitem[{Zhang et~al.(2018)Zhang, Qi, and Manning}]{zhang-etal-2018-graph}
Yuhao Zhang, Peng Qi, and Christopher~D. Manning. 2018.
\newblock Graph convolution over pruned dependency trees improves relation
  extraction.
\newblock In \emph{Proceedings of EMNLP}.

\bibitem[{Zhou(2000)}]{zhou2000block}
Ming Zhou. 2000.
\newblock \href {https://doi.org/10.3115/1117769.1117782} {A block-based robust
  dependency parser for unrestricted {C}hinese text}.
\newblock In \emph{Second {C}hinese Language Processing Workshop}, pages
  78--84, Hong Kong, China. Association for Computational Linguistics.

\bibitem[{Zhu et~al.(2019)Zhu, Li, Zhu, Qian, Zhang, and
  Zhou}]{zhu-etal-2019-modeling}
Jie Zhu, Junhui Li, Muhua Zhu, Longhua Qian, Min Zhang, and Guodong Zhou. 2019.
\newblock Modeling graph structure in transformer for better {AMR}-to-text
  generation.
\newblock In \emph{Proceedings of the 2019 Conference on Empirical Methods in
  Natural Language Processing and the 9th International Joint Conference on
  Natural Language Processing (EMNLP-IJCNLP)}, Hong Kong, China. Association
  for Computational Linguistics.

\end{thebibliography}
\bibliographystyle{acl_natbib}

\end{CJK*}		  
\end{document}